\documentclass[runningheads]{llncs}
\usepackage[T1]{fontenc}
\usepackage{graphicx}
\usepackage{amsmath}
\usepackage{bbm}
\usepackage{multirow,multicol}
\usepackage{subcaption}
\usepackage{booktabs}
\usepackage{fvextra} 
\begin{document}
\title{KIEval: Evaluation Metric\\for Document Key Information Extraction}
\author{Minsoo Khang \and
Sang Chul Jung \and
Sungrae Park \and
Teakgyu Hong}
\institute{Upstage AI, South Korea \\ 
\email{\{mkhang, eric, sungrae.park, tghong\}@upstage.ai}}

\maketitle              %

\begin{abstract}

Document Key Information Extraction (KIE) is a technology that transforms valuable information in document images into structured data, and it has become an essential function in industrial settings.
However, current evaluation metrics of this technology do not accurately reflect the critical attributes of its industrial applications.
In this paper, we present KIEval, a novel application-centric evaluation metric for Document KIE models.
Unlike prior metrics, KIEval assesses Document KIE models not just on the extraction of individual information (entity) but also of the structured information (grouping).
Evaluation of structured information provides assessment of Document KIE models that are more reflective of extracting grouped information from documents in industrial settings.
Designed with industrial application in mind, we believe that KIEval can become a standard evaluation metric for developing or applying Document KIE models in practice.
The code will be publicly available.


\end{abstract}

\section{Introduction}

\begin{figure}[h!]
    \centering
    \includegraphics[width=\linewidth]{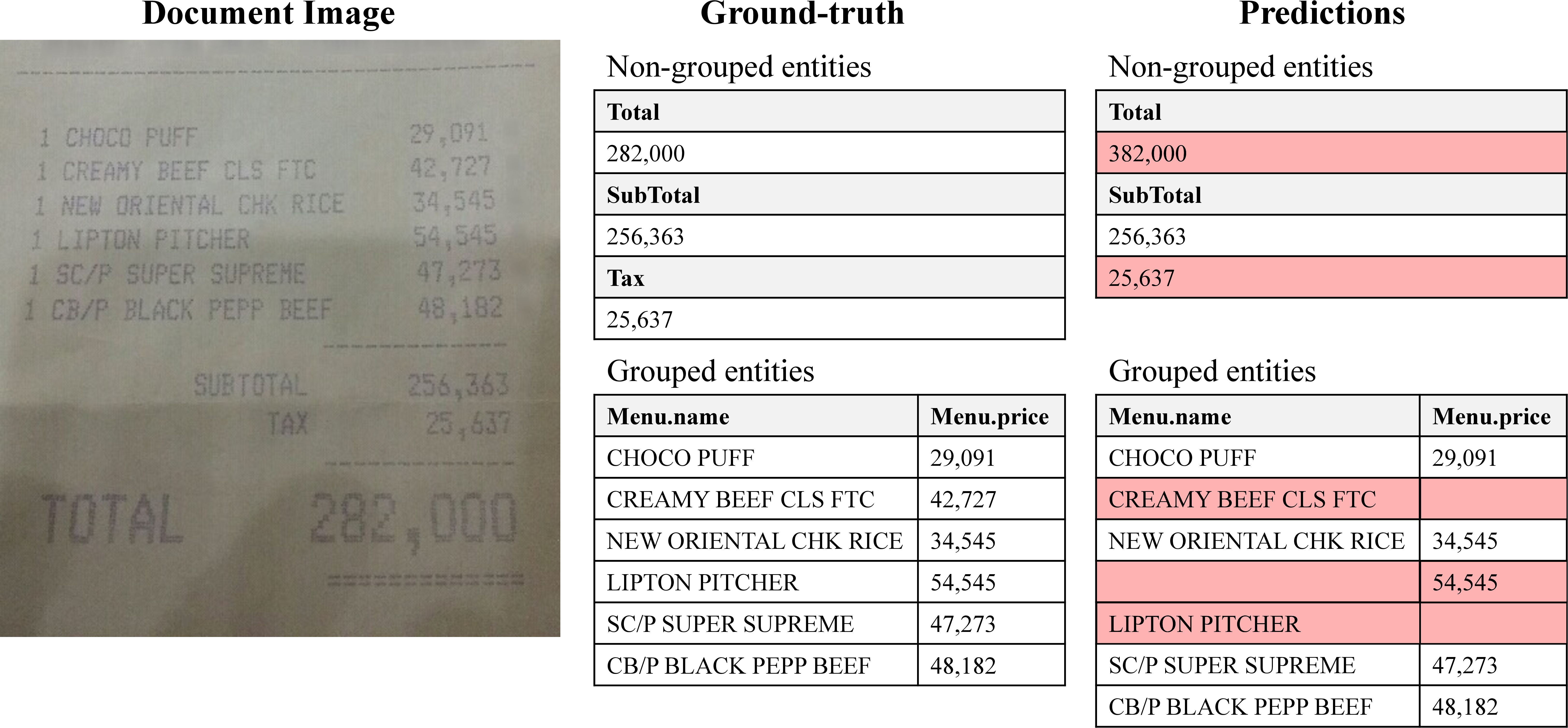}
    \caption{Example of CORD dataset (receipts). The dataset has non-grouped and grouped entities (non-grouped entities form a special group), and requires structured predictions including Menu groups: Menu.name and Menu.price. Errors in model predictions are not limited to individual key-value pair errors but also in the extraction of structural relation between entities (marked in red). Both error types must be considered in Document KIE model evaluation.%
    }
    \label{fig:teaser}
\end{figure}

Document Key Information Extraction (KIE) is a well-known task of converting information from document images into structured data and has gained much attention from both the academia and industry over the years ~\cite{xu2020layoutlm,xu2020layoutlmv2,huang2022layoutlmv3,garncarek2021lambert,hong2022bros,kim2022ocr,peng2022ernie,lee2023pix2struct,hwang2020spatial}. One common application of Document KIE in industrial settings lies in Robotic Process Automation (RPA) of document digitisation which aims to extract, structure, and store the data in document images into databases for various downstream applications. Information extracted from documents is often presented as key-value pairs (e.g. Menu.name: ``AMERICANO'') that are frequently interrelated (e.g. Menu.name \& Menu.price), forming the basis of structured information in documents.

\begin{figure}[t!]
    \centering
    \includegraphics[width=\linewidth]{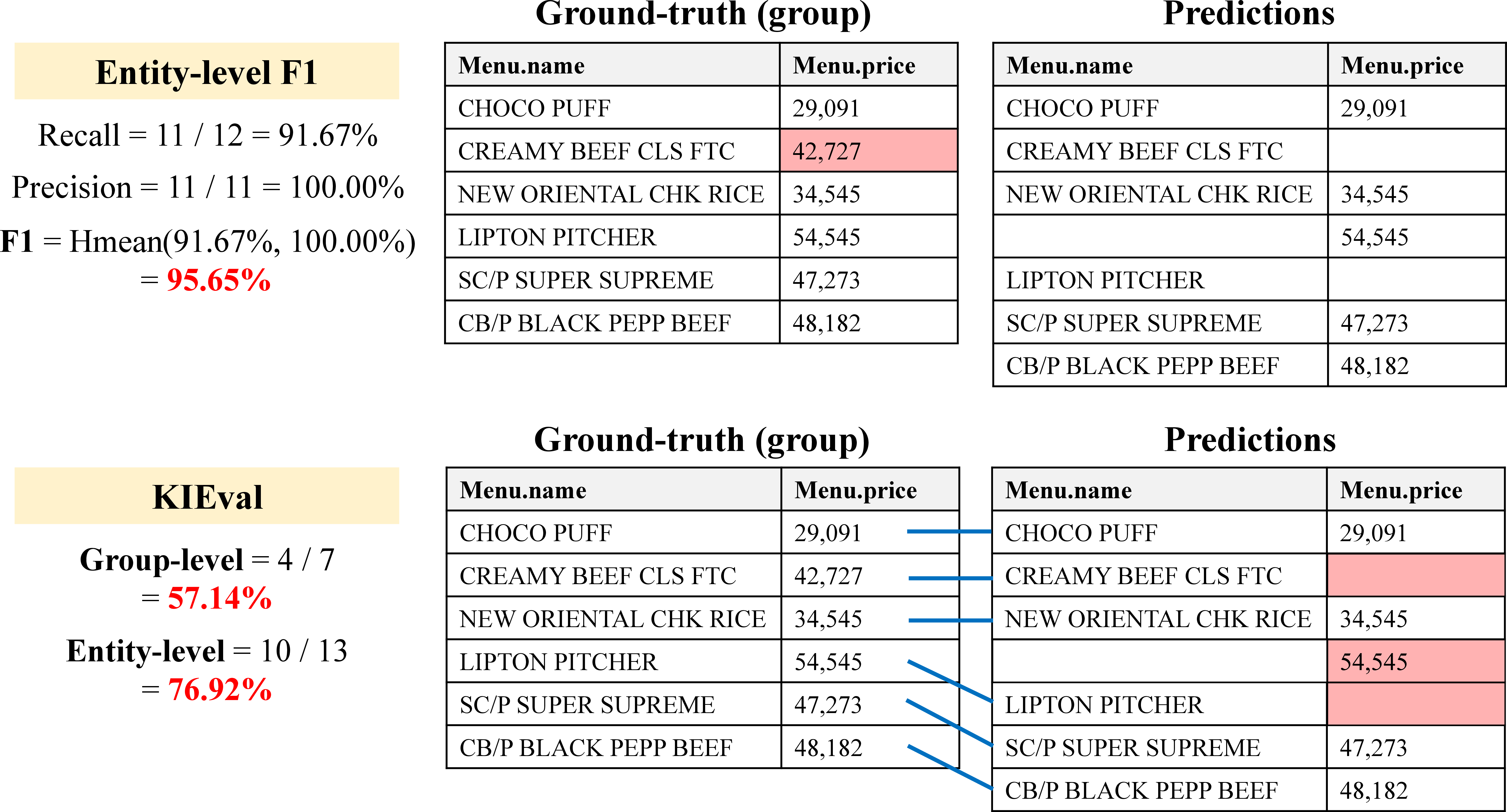}
    \caption{Comparison of evaluation metrics for KIE tasks. Note that the ground-truth and predictions follow the same Document Image in Fig.~\ref{fig:teaser}. The red boxes indicate the errors accounted for by the respective metrics during evaluation. Entity-level F1
    does not account for structural relations unlike the proposed KIEval metric which performs both Entity-level and Group-level evaluations based on the group-matching information (blue links).
    }
    \label{fig:teaser2}
    \vspace{-1em}
\end{figure}

Despite such application settings, a standardized evaluation metric for Document KIE models has yet to be established and existing metrics used in prior works fail to consider several key components from the application's standpoint. The main causes of disparity between the existing evaluation metrics and application settings can be attributed to: neglecting structured nature of information in assessment and insufficient alignment of metric formulation with the industrial applications. Detailed explanations of these causes are as follows (corresponding visualisations are shown in Fig.~\ref{fig:teaser} and~\ref{fig:teaser2}):

\textit{Structured nature in information} refers to the presence of structural relation between key-value pairs in documents. Referring to the example in Fig.~\ref{fig:teaser} and~\ref{fig:teaser2}, each values of the entity-type, Menu.name, has contextual linkage to different values of Menu.price. Existing metrics, however, mainly focus on the assessment of individual entity extraction, while reflecting limited or no evaluation for extraction of such structured information. In industrial applications, however, the lack of such structured information can lead to critical information loss when storing data in relational databases for downstream tasks.

\textit{Insufficient alignment} of existing metrics' design refers to gaps arising due to the formulations that are not fully representative of Document KIE applications in industrial settings. Existing metrics, such as the Entity-level F1 metric, often distinguishes KIE model's erroneous prediction (False-Positive, FP) from missed prediction (False-Negative, FN) in metric formulations. Such distinction, while well-suited for model development, precipitates clear disparity with application settings where KIE errors are often perceived in number of correction counts needed. It is worth noting that, correction count refers to number of value editing (one of substitution, addition, or deletion) steps needed to convert KIE predictions to ground-truth values.

To address the causes of disparity identified above, we propose an evaluation metric with application-centric design named: \textbf{KIEval} (Key Information Extraction evaluation). Firstly, KIEval is formulated to provide KIE model assessment in two different levels: entity-level (individual entities such as Menu.name) and group-level (sets of related entities such as Menu.name, Menu.price). In both levels of evaluation, prediction and ground-truth values are matched by conditioning on information structure (group), facilitating structured-information level assessment of KIE models. Secondly, KIEval formulates KIE errors in terms of the number of substitution, addition, or deletion steps needed. Such formulation, instead of the conventional FP and FN, aims to better represent the eventual cost which KIE errors incur in application settings. %

In this work, our key contributions can be summarized as follows. 1) We propose KIEval (Key Information Extraction evaluation) metric for Document KIE which incorporates structured information assessment in both entity and group-level evaluation. 2) Provision of KIE model evaluation in terms of information correction cost, bridging the disparity of existing metrics with industrial applications. 3) We also showcase a use-case study on how KIEval can be applied in RPA systems, highlighting its differences against existing metrics. %

\section{Related Works}
\subsection{Document KIE Methdology}
Various types of approaches to Document KIE have been proposed over the years. One of the notable earlier works, LayoutLM~\cite{xu2020layoutlm}, was the first to propose a multi-modal framework, incorporating both text and layout modalities in key information extraction from documents. The framework's robust performance with simplistic design of BIO-tagging has motivated many follow-up works such as: BROS~\cite{hong2022bros}, LayoutLMv2~\cite{xu2020layoutlmv2}, LayoutLMv3~\cite{huang2022layoutlmv3}, and ERNIE-Layout~\cite{peng2022ernie}. Alternative forms of KIE with improved representational flexibility such as graph-based~\cite{hwang2020spatial} and text generation~\cite{kim2022ocr,lee2023pix2struct} frameworks were proposed in follow-up works to better capture dependencies between entities and effectively capture structured information from documents. With the rise of LLM applications, recent works, such as ICL-D3IE~\cite{he2303icl} and SAIL~\cite{zhang2024sail}, have leveraged the flexibility of LLMs to tackle Document KIE in zero and few-shot settings.

\subsection{Existing Metrics}

\textbf{Entity-level F1 score} is one of the most commonly used metrics for Document KIE model evaluation. Upon extracting entity-wise key-value pairs, they are matched against the ground-truth key-value pairs, where the predicted pair is considered valid if an exact-match can be found in the ground-truth set. Such exact-match statistics are collated across different entities in the dataset to evaluate the model's entity-level F1 score. Commonly used in prior works~\cite{xu2020layoutlm,xu2020layoutlmv2,huang2022layoutlmv3,peng2022ernie}, entity-level F1 score evaluates the degree to which model's extracted information exactly matches the expected key-value content in the document. 

This metric however, not only disregards structural relation between entities during entity-level F1 evaluation but also does not provide any assessment for group-level information extraction. In industrial applications, entities extracted often form meaningful information when grouped with other entities that are structurally related (i.e group), such as the grouping of Menu.name, Menu.quantity, and Menu.price, in receipts. Variations of entity-level F1 score were employed in prior works such as group-constrained entity-level F1 in SPADE ~\cite{hwang2020spatial}, Entity Extraction and Linking F1 in BROS~\cite{hong2022bros}, offering a more comprehensive assessment of KIE models. These variations of entity-level F1 metric underscores the necessity for a standardized metric for KIE model evaluation in the field of Document AI. Furthermore, the absence of direct assessment for group-level information extraction in these metrics highlights the disparity in meeting the industrial application requirements.

\textbf{Tree Edit Distance} (TED) score is another type of KIE evaluation metric, commonly adopted in text-generation based KIE models~\cite{kim2022ocr}. In contrary to the exact-match based entity-level F1 score, TED adopts a soft-match approach to avoid over-penalisation of model's KIE. Edit distance based metric could provide a more objective assessment of the model by mitigating the impact of minor discrepancies, such as those between ``ice cream'' and ``ice-cream'', which could lead to underestimation of model's KIE performance. As discussed in Donut~\cite{kim2022ocr}, TED metric can be applied to KIE models by first representing the prediction and ground truth as trees, before evaluating the edit distance between them. With the structural relation between entities captured using tree representation, this metric offers assessment of not only the model's entity-level KIE performance but also at the group-level.

Despite such capacity of TED metric, its soft-match approach could exacerbate the discrepancy when applied to industrial settings. Taking automatic KIE setting as an example, pairs of information with minor edit distance could refer to completely different items such as ``Pear'' and ``Pea'' or ``7000'' and ``1000''. Consequently, it is required of evaluation metrics to be stringent and provision of partial scores (with edit distance) could offer a misleading KIE assessment.

\textbf{Other} notable metrics include Average Normalized Levenshtein Similarity (ANLS)~\cite{biten2019icdar,tito2021icdar} and hybrid metric of exact-match and edit distance~\cite{yu2023icdar}. ANLS aims to reduce the effect of overestimation of KIE models by constraining the maximum edit-distance between prediction and ground-truth to a pre-specified threshold value (e.g. 0.5) beyond which, no partial score is given. Hybrid metric, on the other hand, is a weighted arithmetic mean of the KIE model's entity-level F1 and inverse Normalized Edit Distance (NED). Nevertheless, these metrics still share the limitations of Entity-level F1 and TED metrics, and do not provide group-level assessment of KIE models.

\section{Problem Definition}

\subsection{Document KIE Task}

Document KIE is a task in the field of Document Understanding (DU), with the objective of extracting structured key-value pairs from Document images. Commonly positioned as the task preceding various knowledge-application operations (e.g. financial data analysis), Document KIE is often faced with two principal challenges: (1) accurate extraction of key-value information, requiring value extraction and entity-key classification with minimal error, and (2) the discernment of structural relations between different key-value pairs, which demands accurate identification of contextual links between key-value pairs, to form a coherent group-level information unit.

\subsection{Measuring Document KIE Model in Industrial Settings}
To create an application-centric evaluation metric, following principal challenges of prior metrics must to be addressed: (1) absence of structural relation in metrics and (2) insufficient alignment of metric formulation with application settings.

Structural relation refers to the contextual linkage between key-value pairs associated with entities in documents. Entity key-value pairs that share contextual linkage are formally defined as a \textbf{group}, such as Menu.name and Menu.price in the CORD example (Fig.~\ref{fig:teaser}). Such structural relations present in documents need to be considered in both entity-level and group-level evaluations. Prior metrics in entity-level evaluation (e.g. Entity-level F1) show limited or no inclusion of structural relation, by evaluating each extracted key-value independent from the remaining key-value pairs. Prediction results visualized in Fig.~\ref{fig:teaser2} clearly demonstrates this where, the prediction of \{Menu.price: ``54,545''\} is not matched with the corresponding \{Menu.name: ``LIPTON PITCHER''\}. Despite predicting a valid Menu.price key-value pair, this prediction is regarded erroneous due to failure in capturing the relation with contextually linked Menu.name key-value pair. This can be intuitively understood as: Menu.price key-value standalone does not provide meaningful information from the application point-of-view, unless paired with the corresponding Menu.name. In group-level evaluations, a more direct assessment of structural relation is conducted where, each group (instead of key-value pair) is treated as a unit of information extracted. Group-level evaluation provides essential assessment of the KIE model especially in industrial applications where information applications are often conducted in contextually related groups (e.g. Menu.name, Menu.price).

\newcommand{\rulesep}{\unskip\ \vrule\ }

\begin{figure}[t]
    \centering
        \begin{subfigure}{0.32\textwidth}
        \includegraphics[width=\linewidth]{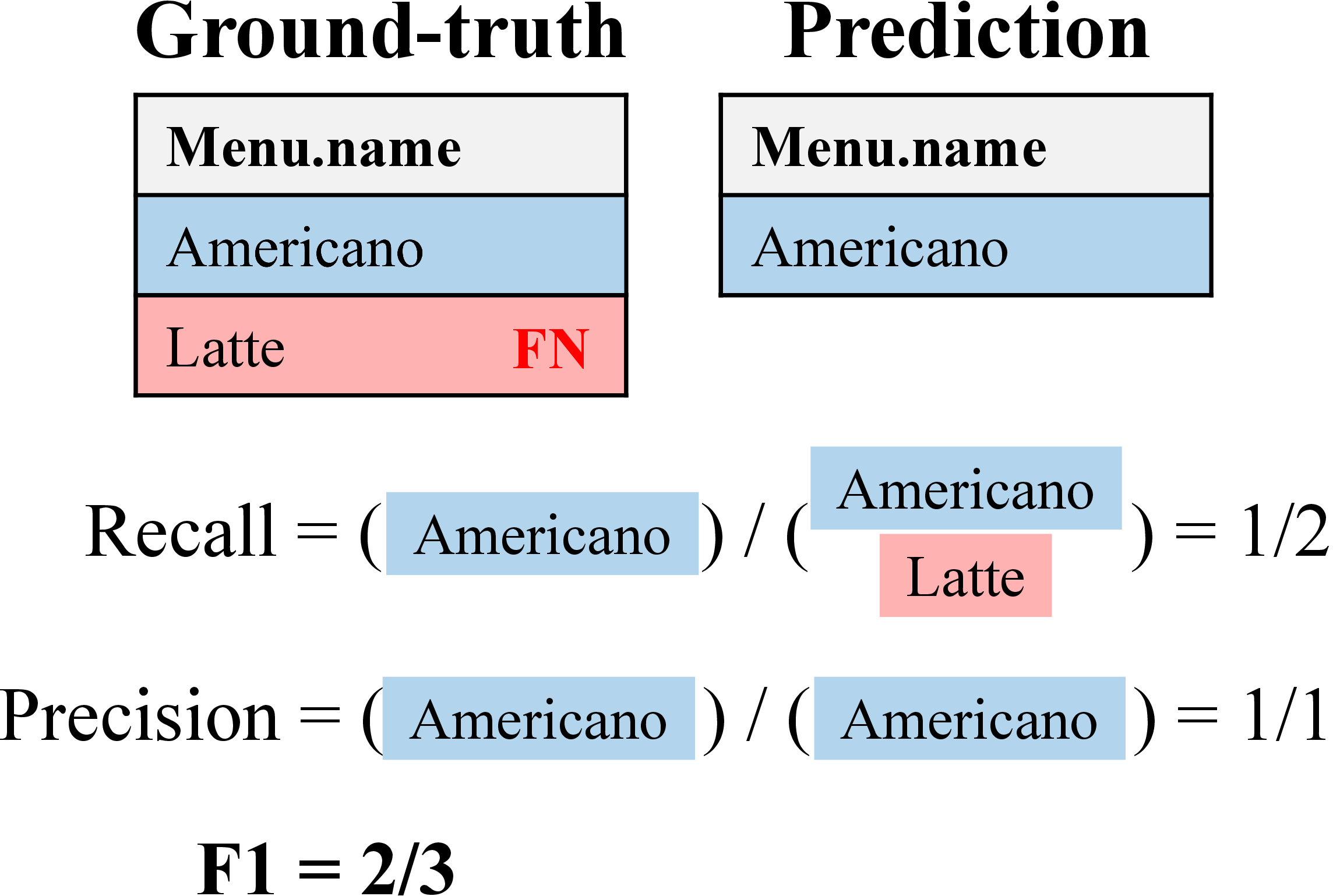}
        \caption{Scenario \#1}
    \end{subfigure}
    \rulesep
    \begin{subfigure}{0.32\textwidth}
        \includegraphics[width=\linewidth]{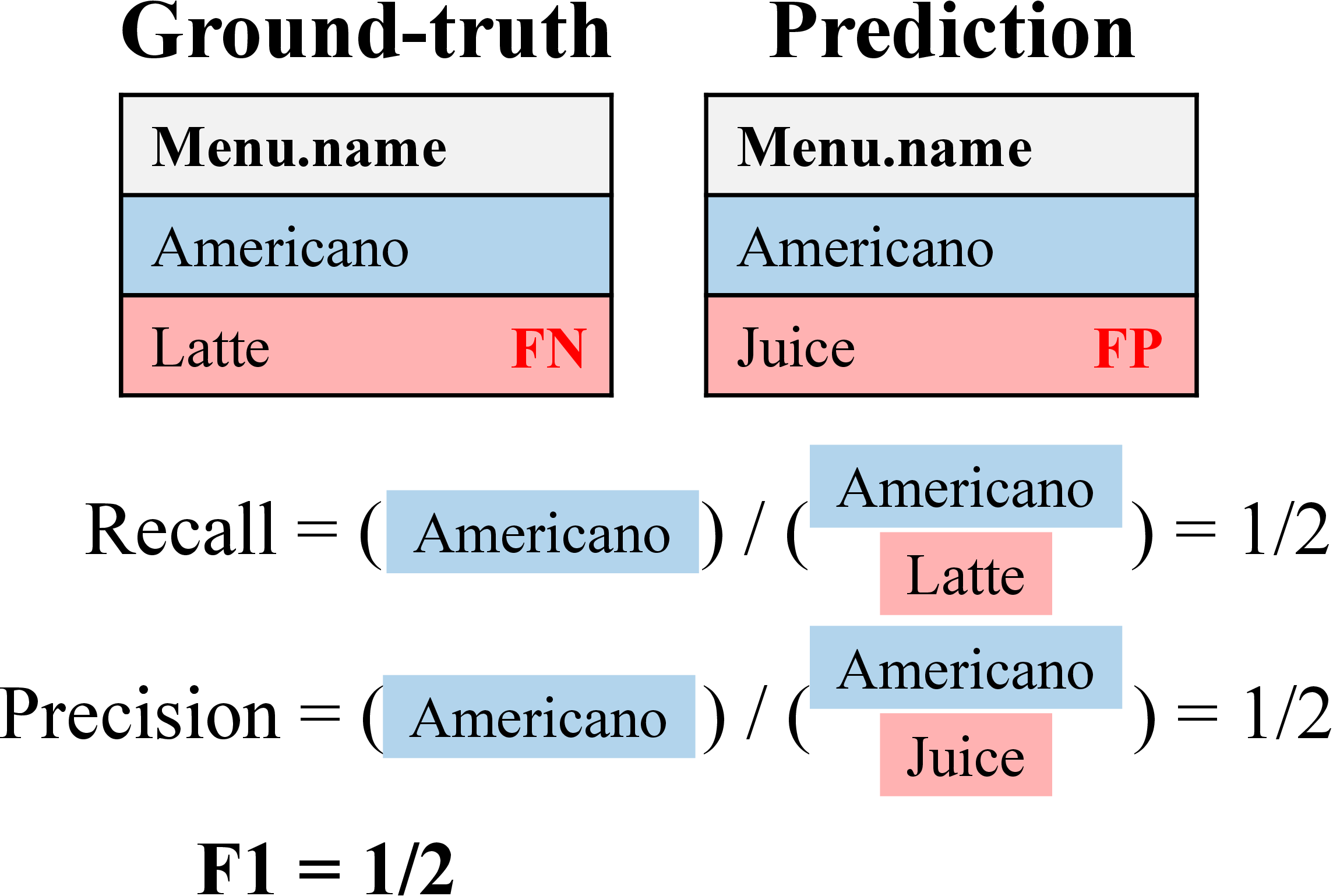}
        \caption{Scenario \#2}
    \end{subfigure}
    \rulesep
    \begin{subfigure}{0.32\textwidth}
        \includegraphics[width=\linewidth]{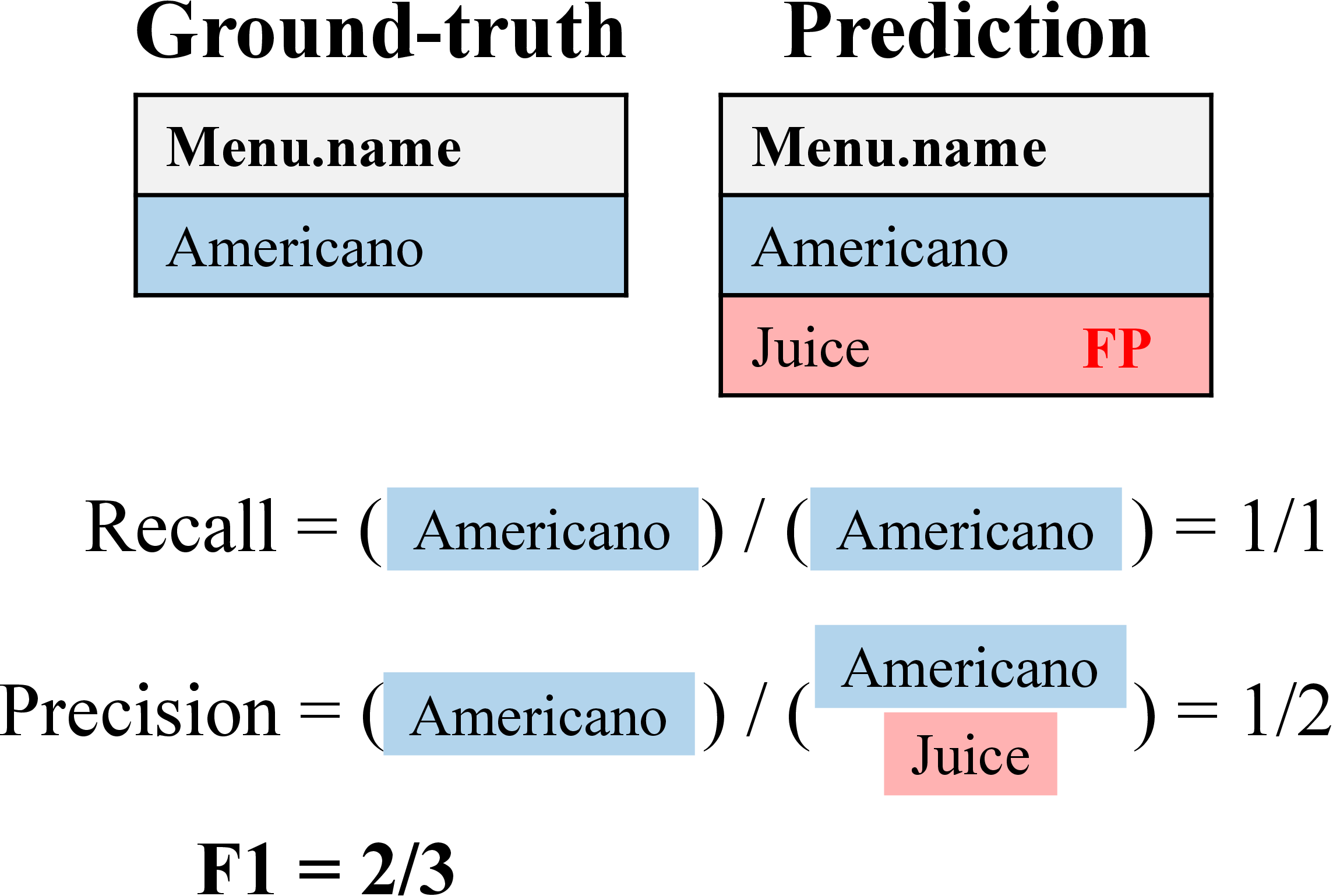}
        \caption{Scenario \#3}
    \end{subfigure}
    \caption{F1 score examples in three different scenarios. From the application perspective, the three different scenarios require the same number of error corrections; (\#1) filling missing information, (\#2) replacing wrong information, and (\#3) deleting the unexpected information. However, in the view of F1 scores, false negative (FN) and false positive (FP) are separately counted to identify representative score value, F1.}
    \label{fig:problem_definition_FPFN}
\end{figure}

In addition to structural relation between key-value pairs, metric formulation is another factor that creates the disparity between the (prior works') model-centric and (KIEval's) application-centric design. Model-centric metric formulations often distinguish model's erroneous prediction (FP) from missed prediction (FN), such as in Entity-level F1 metric. In industrial applications, however, it is more relevant to assess KIE models in terms of additional cost incurred due to KIE errors. To elaborate, with reference to Fig.~\ref{fig:problem_definition_FPFN}, Entity-level F1 evaluation across the three scenarios implies lower KIE performance in scenario 2. From the application perspective, however, all of the three scenarios' predictions incur the same cost of one editing operation (addition, substitution, or deletion) in KIE automation. Consequently, it is imperative to develop an application-centric metric formulation that accurately reflects the actual application settings.

Based on the key challenges of application-centric design defined above, our work's proposed metric, KIEval, is designed with these factors in mind to bridge the disparity between the current metrics and industrial applications.

\section{KIEval}

\subsection{Structured Evaluation -- Entity and Group Level}
In KIEval, to integrate structural relation into the KIE evaluation, \textit{group-matching} was conducted between the predicted and ground-truth key-value pairs prior to entity-level and group-level evaluations. While variant of \textit{group-matching} for entity-level evaluation was employed in~\cite{hwang2020spatial}, the lack of formal definition underscores its significance in the view of KIE metric standardisation. 
To illustrate, let $\mathbf{PR} = \{{\text{pr}}_1, {\text{pr}}_2, ..., {\text{pr}}_{N}\}$ be a set of predicted groups and $\mathbf{GT} = \{\text{gt}_1, \text{gt}_2, ..., {\text{gt}}_{M}\}$ be ground-truth groups, where each group consists of a set of entities represented by tuples, (\text{entity-type}, \text{value}). The non-group entities (i.e. company.name and company.number in receipt) are included in $1$-st group to represent all entities with the same structural format. For the formal definition of \textit{group-matching}, we define a matching score $S_{(n,m)\textbf{}}$ counting the identical entities between ${\text{pr}}_n$ and ${\text{gt}}_m$. Based on the matching scores between groups, each prediction group is matched with a ground-truth group through Hungarian matching to obtain a group-matched set of groups, $\mathbf{G} = \{(n_1, m_1), (n_2, m_2), ..., \}$, where $n_g$ and $m_g$ indicate the $g$-th matched indices of predicted and ground-truth groups, respectively, and $|\mathbf{G}|$ results as $\min(N, M)$. The group-matching can be defined as follows:
\begin{equation}
    \label{eqn:groupmatchHungarian}
    \mathbf{G} = \text{Hungarian}(\mathbf{PR}, \mathbf{GT}, \mathbf{S})
\end{equation}
\noindent where $\mathbf{S}$ indicates a set of matching scores, $S_{(n,m)}$, between all pairs between predictions and ground-truth. For an entity $e$ at a matching $(n,m)$, F1 statistics such as True-Positive (\textit{TP}), False-Negative (\textit{FN}), and False-Positive (\textit{FP}) can be calculated as follows;
\begin{equation}
    \label{eqn:entity_F1_statistics}
    {TP}_{(n,m)}^{e} = S_{(n,m)}^{e},
    {FN}_{(n,m)}^{e} = \text{N}_{e}({\text{gt}}_m)-S_{(n,m)}^{e},
    {FP}_{(n,m)}^{e} = \text{N}_{e}({\text{pr}}_n)-S_{(n,m)}^{e} 
\end{equation}
where $S_{(n,m)}^{e}$ indicates the number of identical entity pairs, which has entity-type $e$ between $n$-th predicted and $m$-th ground-truth groups. The $\text{N}_{e}(\cdot)$ represents the operation counting entity-type $e$ in a group. In other words, ${TP}_{(n,m)}^{e}$ indicates the matched entity, and ${FN}_{(n,m)}^{e}$ and ${FP}_{(n,m)}^{e}$ represent the remaining ground-truths and predictions in the specific match $(n,m)$ in terms of entity type $e$, respectively. To calculate a final cumulated score, \textit{KIEval Entity F1}, the total F1 statistics are identified as follows:
\begin{align}
    \label{eqn:total_entity_F1_statistics}
    {TP}^{\text{entity}} & = \sum_{(n,m)\in \mathbf{G}} \sum_{e} {TP}_{(n,m)}^{e} \\
    {FN}^{\text{entity}} & = \sum_{m}^{M} \sum_{e} \text{N}_{e}({\text{gt}}_m) - TP^{\text{entity}} \\
    {FP}^{\text{entity}} & = \sum_{n}^{N} \sum_{e} \text{N}_{e}({\text{pr}}_n) -TP^{\text{entity}} 
\end{align}
The statistics can be used to calculate \textit{KIEval Entity F1} metric using standard precision and recall manners.

Group-level evaluation, \textit{KIEval Group F1}, is also conducted on the group-matched $\mathbf{G}$ where F1 statistics are evaluated across different groups. Unlike \textit{KIEval Entity F1} which treats all entities in a group as a unit of information to evaluate F1 statistics, \textit{KIEval Group F1} evaluates on the entire group as a unit of information. It should be noted that, group-level evaluation is conducted on all but the first element of $\mathbf{G}$ (i.e. $\mathbf{G}'$) as the first element represent non-group entities.
\begin{equation}
    \label{eqn:onlyGroupedG}
    \mathbf{G}'=\mathbf{G} \setminus {(n_1, m_1)}
\end{equation}
\begin{equation}
    \label{eqn:groupingTP}
    {TP}^{\text{group}} = \sum_{(n,m)\in\mathbf{G}'} \mathbbm{1}[ S_{(n,m)}^e=\text{N}_{e}({\text{gt}}_m)=\text{N}_{e}({\text{pr}}_n) \;\;\;\forall e]
\end{equation}
\noindent Eq.~\ref{eqn:groupingTP} shows formulation of group-level True-Positive measure where counting identical pairs of prediction and ground truth groups in $\mathbf{G}'$. In the equation, $\mathbbm{1}[\cdot]$ indicates a binary operator providing 1 when the predicted and ground-truth groups are identical. FN and FP are calculated by counting the remaining ground-truth and predicted groups, respectively. Finally, \textit{KIEval Group F1} can be identified with the same precision and recall fashion. Based on the formal definition of \textit{KIEval Entity F1} and \textit{KIEval Group F1} above, both formulations aim to incorporate structure relation assessment in evaluation at the entity and group-level, respectively.

\subsection{Aligned Metric Formulation}
While distinction of model's erroneous prediction and missed prediction as FP and FN in metric formulation could be well-suited from the model-centric point-of-view, its misalignment in the standpoint of industrial application has motivated the formulation of our metric. KIEval's application-centric design addresses this misalignment by conceptualizing KIE errors as correction costs incurred in application settings. \textit{Correction} refers to one of the three editing steps: substitution, addition, and deletion of prediction values to match the ground-truth. For an entity $e$ at the matching condition $(n,m) \in \mathbf{G}$, the steps can be defined in terms of FN and FP as follows:
\begin{align}
    \label{eqn:CorrectionCost}
    \text{Subs}_{(n,m)}^{e} & = \min(FP_{(n,m)}^{e},FN_{(n,m)}^{e}) \\
    \text{Add}_{(n,m)}^{e} & = FN_{(n,m)}^{e} - \text{Subs}_{(n,m)}^{e} \\
    \text{Del}_{(n,m)}^{e} & = FP_{(n,m)}^{e} - \text{Subs}_{(n,m)}^{e}
\end{align}
As can be seen, the substitution is defined as the minimum number of $FP_{(n,m)}^{e}$ and $FN_{(n,m)}^{e}$, which indicates the number of predictions that require modifications to match corresponding ground-truth values. The addition and deletion are the number of remaining $FN_{(n,m)}^{e}$ and $FP_{(n,m)}^{e}$, respectively. The number of error, $\text{Error}_{(n,m)}^{e} = \text{Subs}_{(n,m)}^{e} + \text{Add}_{(n,m)}^{e} + \text{Del}_{(n,m)}^{e}$, is represented by summing the three error corrections. The total number of error can be defined as follows;
\begin{equation}
    \text{Error} = \sum_{(n,m)\in\mathbf{G}} \sum_{e} \text{Error}_{(n,m)}^{e} + \underbrace{\sum_{(*,m)\notin\mathbf{G}}\sum_{e} \text{N}_{e}({\text{gt}}_m)}_{\text{Add unmatched gt}} + \underbrace{\sum_{(n,*)\notin\mathbf{G}}\sum_{e} \text{N}_{e}({\text{pr}}_n)}_{\text{Del unmatched pr}}
\end{equation}
Here, the first term on the right-hand side indicates the number of error corrections in the group match condition $\mathbf{G}$, and the second and third terms represent the number of additions and deletions, respectively, for the non-matched groups. Finally, $\text{KIEval}_{\text{Aligned}}$ is calculated with the $\text{Error}$ and the number of correct values, $TP$. The following equation shows the formulation;
\begin{equation}
    \text{KIEval}_{\text{Aligned}} = \frac{TP^{\text{entity}}}{TP^{\text{entity}} +  \text{Error}}
\end{equation}
The $\text{KIEval}_{\text{Aligned}}$ not only better aligns with industrial applications, but also benefits from high interpretability due to its formulations in terms of well-known F1 components: TP, FP, and FN. %

\section{Experiment Settings}

\subsection{Datasets}
\begin{figure}[t]
    \centering
    \includegraphics[width=\linewidth]{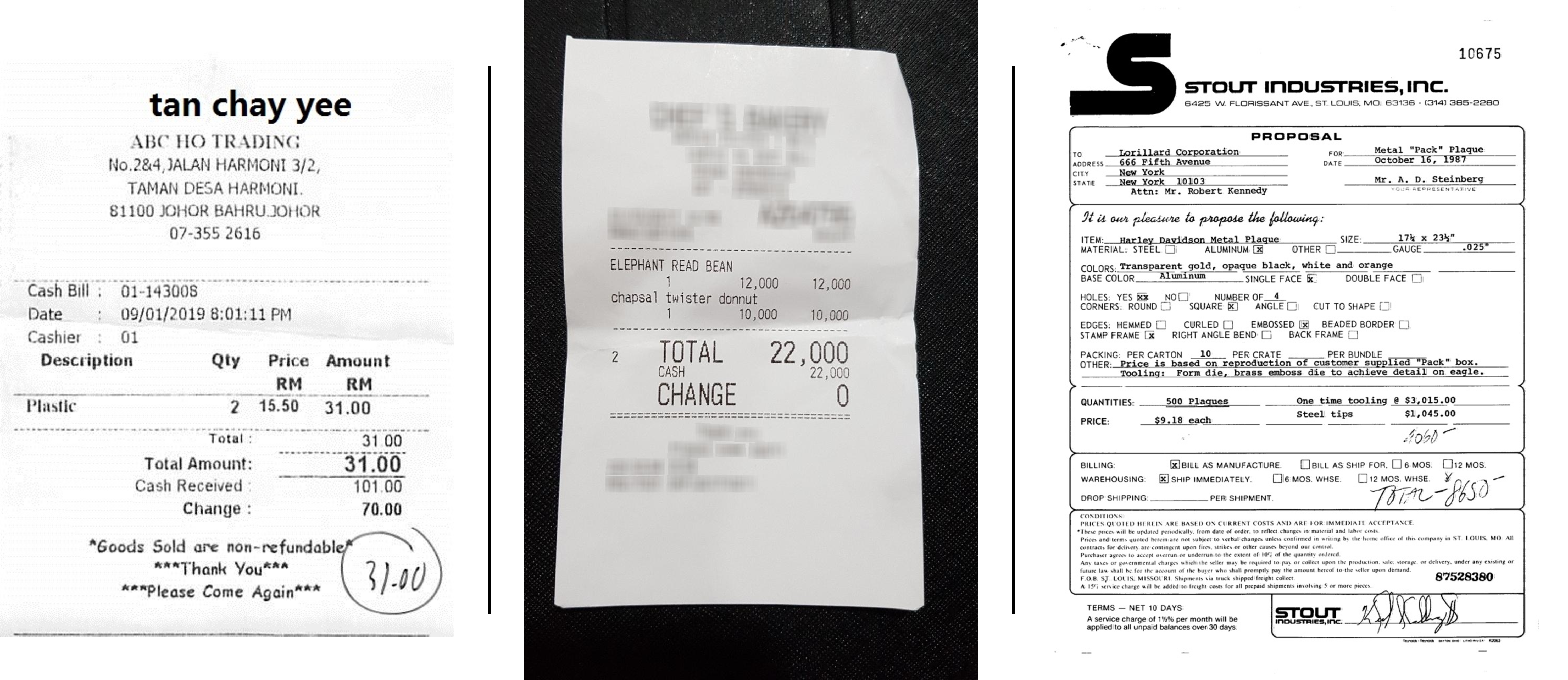}
    \caption{Sample images from the SROIE (left), CORD (center), and FUNSD (right) datasets.}
    \label{fig:datasets}
\end{figure}

Experiments were conducted with the KIEval metric on models trained using three widely used benchmark datasets in the Document KIE task, namely: SROIE, CORD and FUNSD, shown in Fig~\ref{fig:datasets}.

\textbf{SROIE} dataset refers to the dataset introduced in task 3 of \textit{Scanned receipts OCR and information extraction} challenge of ICDAR 2019\footnote{https://rrc.cvc.uab.es/?ch=13}. This dataset comprises of 626 train and 347 test receipt images, requiring participants' models to extract key-value pairs of 4 entities: Company, Date, Address and Total price from these images.

Consolidated Receipt Dataset (\textbf{CORD})~\cite{park2019cord} comprises of receipt images from shops and restaurants designed for the task of extracting grouped entities. Dataset's annotation consists of 30 entities, which are categorized into 4 groups: Menu, Void menu, Subtotal, and Total. Entities within each group are contextually linked such as: Menu.name, Menu.price and Menu.quantity. There are 800 training, 100 validation and 100 testing images.

Form Understanding in Noisy Scanned Documents (\textbf{FUNSD})~\cite{jaume2019funsd} consists of 149 training and 50 test documents, which are noisy, scanned, and have various layouts. This dataset is composed of three entities: Header, Question, and Answer. FUNSD, unlike aforementioned datasets, allows each entity to hold multiple values within the same document image. To maintain consistency with prior works on FUNSD, all entities are regarded as non-group in our experiments.

\subsection{Document KIE Models}

Current works on Document KIE can be largely categories into two frameworks: sequence labeling and generative frameworks.

Prior works in the sequence labeling framework adopt tagging-based approach to extract key-value pairs from document images. In detail, with reference to CORD sample image in Fig.~\ref{fig:datasets}, OCR is first applied to extract texts such as ``Vt Pep Mocha'' before tokenizing it into ``Vt'', ``Pep'' and ``Mocha''. The KIE model then processes these tokens, often conditioned with layout and image information, to provide token-wise label (e.g. BIO tag) classifications such as ``B-Menu.name'', ``I-Menu.name'', and ``I-Menu.name'' to the example text respectively. Tokenized texts along with their corresponding token-level tags are then postprocessed to form the final key-value pairs (e.g. Menu.name: ``Vt Pep Mocha''). Representative works in this framework include: LayoutLM family~\cite{xu2020layoutlm,xu2020layoutlmv2,xu2022xfund,huang2022layoutlmv3}, StructuralLM~\cite{li2021structurallm}, BROS~\cite{hong2022bros}, LiLT~\cite{wang2022lilt}, and DocFormer~\cite{appalaraju2021docformer}.

Generative framework based models conduct KIE from document images by directly generating the key-value pairs as text. Taking the same CORD example in Fig.~\ref{fig:datasets}, generative KIE models generates text sequence of key-value pairs such as: \{Menu.name: ``Vt Pep Mocha'', Menu.price: ``4.95''\}. OCR information can also be provided as auxiliary input to these KIE models. Notable generation methods include TILT~\cite{powalski2021going}, Donut~\cite{kim2022ocr}, and Pix2Struct~\cite{lee2023pix2struct}, where ResNet~\cite{he2016deep} or ViT~\cite{dosovitskiy2020image} is commonly used for image encoder and Transformer decoder~\cite{vaswani2017attention} for text decoder.

In this work, we conduct experiments using LayoutXLM~\cite{xu2022xfund} and LayoutLMv3~\cite{huang2022layoutlmv3} models for the sequence labeling framework, and the Donut~\cite{kim2022ocr} model for the generative framework. Given recent advancements in large language model (LLM) applications for document intelligence, we also conduct zero-shot LLM-based KIE experiments with GPT-4o~\cite{openai2023gpt-4v}, Qwen2-VL~\cite{Qwen2VL} and InternVL 2.5~\cite{chen2024expanding}. These experiments demonstrate how KIEval can provide additional insights into LLM evaluation within the KIE context.

\subsection{Grouping Information}

With prior KIE models mainly designed for KIE at the entity-level, we adopt simple methodology to extract grouping information prior to KIEval evaluations.

For models of sequence labeling framework, a simple slot filling method is adopted for grouping. To elaborate, given the set of entity-types constituting a group (e.g. Menu.name, Menu.price, ... in CORD's Menu group), KIE model's output of these entity-types are sequentially filled in a slot filling manner to form groups. While different approaches for grouping extraction can be adopted, such as relation extraction~\cite{xu2022xfund} or graph-based method~\cite{hwang2020spatial} on top of the KIE models, for the purpose of assessing the effectiveness of KIEval metric, a simple grouping method was employed. For text-generation based models, group-level information can be extracted by simple structuring of the target key-value pair text sequence such as JSON format strings.

\subsection{Experiment Details}

All sequence labeling and generation models were trained for 1,000 steps with a batch size of 16.
The initial learning rate was set to 5e-5, along with linear learning rate decay.
We used the provided OCR annotations along with images for experiments involving LayoutXLM~\cite{xu2022xfund} and LayoutLMv3~\cite{huang2022layoutlmv3} while only the document image was provided for Donut~\cite{kim2022ocr} experiments. For reproducibility, all experiments were conducted using the models and datasets uploaded to Hugging Face Models and Datasets\footnote{https://huggingface.co}.
Details can be found in Appendix A.
For multimodal LLMs, all experiments were conducted in zero-shot setting, and the prompts used can be found in Appendix B.

\section{Results and Discussion}

\subsection{Structured Evaluation}

\begin{figure}[t]
    \centering
    \includegraphics[width=\linewidth]{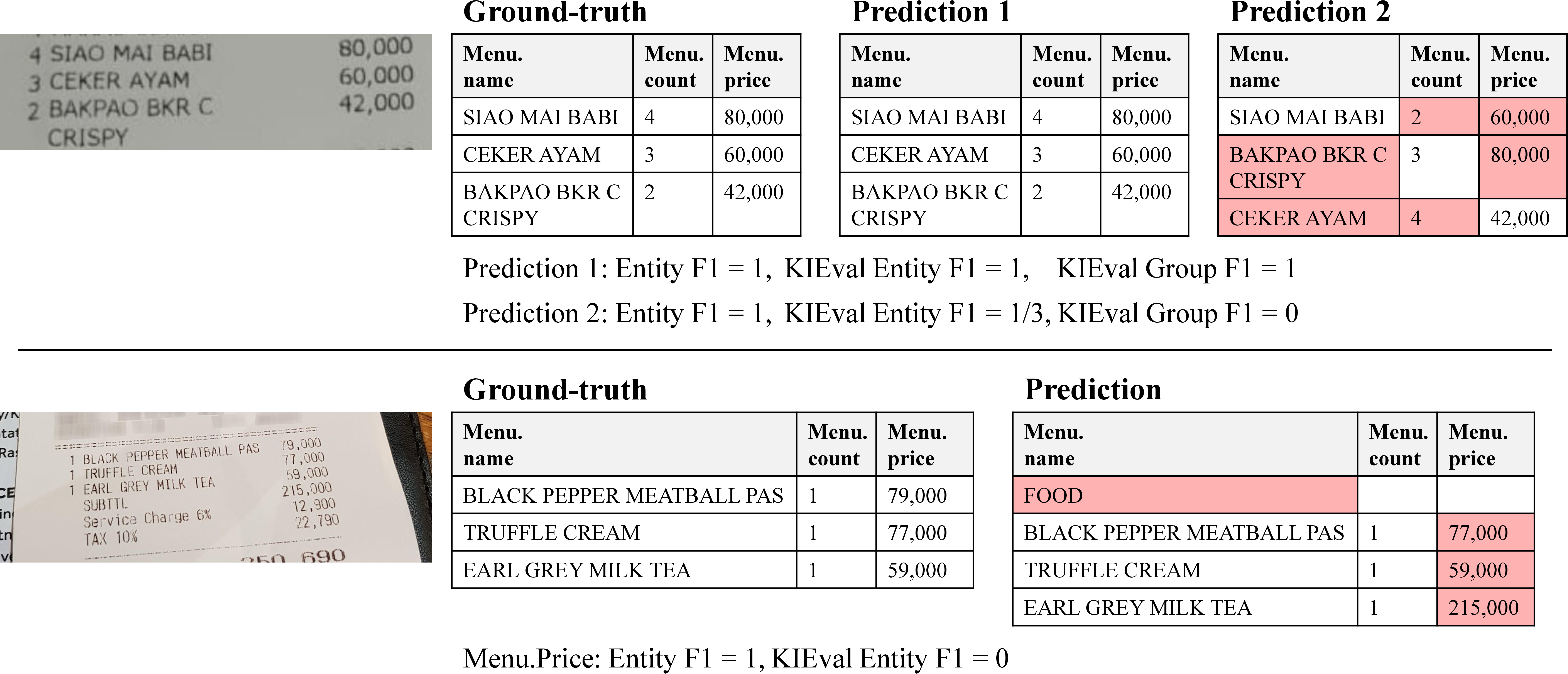}
    \caption{Examples illustrating the difference between Entity F1 and KIEval. The above scenario is constructed to showcase metric disparities, whereas the scenario below is based on real prediction result from the Donut model.}
    \label{fig:pitfalls_entity_f1_merged}
\end{figure}

The conventional Entity F1 metric fails to accurately represent the KIE model's performance due to absence of structural relation consideration. Fig.~\ref{fig:pitfalls_entity_f1_merged}(top) illustrates conceptual examples with corresponding metric scores across Entity F1, KIEval Entity F1 and KIEval Group F1. In Fig.~\ref{fig:pitfalls_entity_f1_merged}(top), while both Prediction 1 and 2 display accurate entity-level key-value pair extractions, contextual relations (grouping) between different key-value pairs are not well-captured in Prediction 2. Such observations are not well-reflected in the conventional Entity F1 metric, scoring 1.0 across both predictions. 

In industrial applications where both the key-value and contextual linkage information need to be extracted (if present), Entity F1's insensitivity towards the latter could lead to sub-optimal reflection of the KIE model's performance especially in RPA applications. KIEval Entity F1 and KIEval Group F1, on the contrary, provide distinct evaluations across the two predictions by taking into account of structural relations in the formulations. In KIEval Entity F1, despite error-free extraction of key-value pairs for each entity-type, Prediction 2 is penalized for its grouping errors, resulting in a score of $1/3$. Similarly in KIEval Group F1, where each group is treated as a single-unit of information instead of key-value pairs, Prediction 2 is evaluated to be completely incorrect, which is not discernible from the Entity F1 metric.

Fig.~\ref{fig:pitfalls_entity_f1_merged}(bottom) depicts a sampled inference result of the Donut (generation KIE) model. Despite accurate extraction of Menu.name key-values, contextual linkage with other entity types are misaligned possibly due to tilt rotation of the receipt image. The conventional Entity F1 score of Menu.price entity does not reflect this error and assigns a full score of 1.0 unlike KIEval Entity F1 which penalizes the prediction accordingly.

\begin{table}[t]
\centering
\begin{tabular}{lcccccccc}
\toprule
\multirow{2}{*}{} & \multicolumn{3}{c}{LayoutXLM} & \multicolumn{3}{c}{LayoutLMv3} & \multicolumn{2}{c}{Donut} \\
                  & SROIE    & CORD     & FUNSD   & SROIE     & CORD     & FUNSD   & SROIE       & CORD        \\

\midrule
Entity F1         & 91.77   & 95.43   & 84.02       & 91.87    & 95.13   & 85.87       & 83.85      & 84.93      \\
nTED              & 97.24   & 94.86   & 61.96       & 96.91    & 94.43   & 69.36       & 96.17      & 90.62      \\
\midrule
KIEval Entity F1            & 91.77   & 92.88   & 84.02       & 91.87    & 91.84   & 85.87       & 83.85      & 84.47     \\
KIEval Group F1            & -   & 82.68   & -       & -    & 82.11   & -       & -      & 68.26     \\
KIEval$_{\text{Aligned}}$            & 90.32   & 89.02   & 79.22       & 91.15    & 88.15   & 80.22       & 83.57      & 79.70     \\
\bottomrule

\end{tabular}
\caption{Comparision of Entity F1, nTED, and KIEval. When group entities are absent, Entity F1 and KIEval Entity F1 yield identical values. Note: Donut displays substantially lower performance than other models due to its sole reliance on image input, unlike other models' use of ground-truth OCR annotations.}
\label{tbl:all_scores}
\end{table}

Evaluation results for different metrics across all models and datasets are shown in Table~\ref{tbl:all_scores}. For nTED, its soft-match approach inaccurately compares KIE performance, as seen in LayoutXLM and LayoutLMv3 on SROIE, where trends differ from Entity F1 and KIEval Entity F1. For Entity F1, the differences compared to KIEval Entity F1 are prominent in CORD dataset where contextual links (grouping) between entities are present. KIEval Entity F1 consistently underperforms compared to Entity F1 in CORD across all models, despite achieving equivalent scores in SROIE and FUNSD. This discrepancy highlights the overestimation of KIE model performance when structural relations are ignored, while the metric converges to Entity F1 in datasets without grouping.

\begin{table}[t]
\centering
\begin{tabular}{lcccc}
\toprule
\multirow{1}{*}{} & \multicolumn{1}{c}{Donut} & \multicolumn{1}{c}{GPT-4o} & \multicolumn{1}{c}{Qwen2-VL} & \multicolumn{1}{c}{InternVL 2.5} \\

\midrule
Entity F1         & 84.93   & 73.56   & 77.07       & 54.99  \\
KIEval Entity F1            & 84.47   & 72.93   & 77.07       & 54.54 \\
\midrule
Difference & 0.46 & 0.63 & 0.00 & 0.45 \\
\bottomrule

\end{tabular}
\caption{Comparison of Entity F1 and KIEval Entity F1 across generative models including multimodal LLMs on the CORD dataset. All multimodal LLMs are evaluated in a zero-shot setting. The difference between Entity F1 and KIEval Entity F1 serves to highlight information structure awareness of LLMs in KIE.}
\label{tbl:llm_kieval}
\end{table}

\begin{figure}[t]
    \centering
    \includegraphics[width=\linewidth]{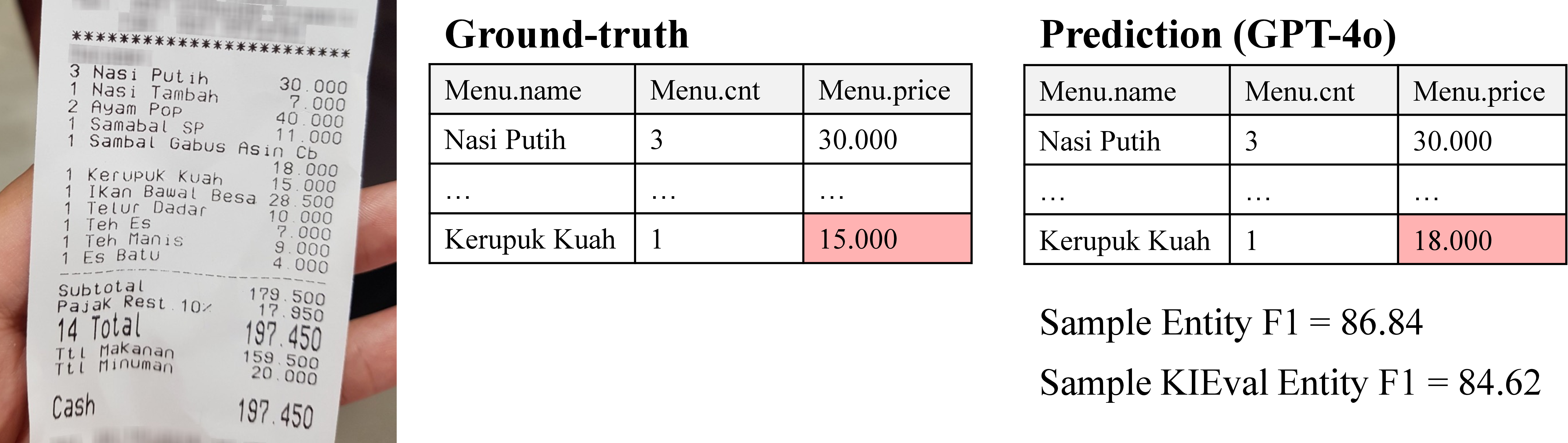}
    \caption{Sample of CORD dataset, illustrating the performance gap between Entity F1 and KIEval Entity F1 in GPT-4o, highlighting the importance of structure awareness evaluation on top of the existing KIE metric.}
    \label{fig:LLM_KIEval}
\end{figure}

The discrepancy is also evident in multimodal LLM evaluation, as shown in Table~\ref{tbl:llm_kieval}. The difference between Entity F1 and KIEval Entity F1 provides deeper insight into the LLM's ability in grouping correctly extracted information into the expected semantic structures. Based on the CORD results, Qwen2VL outperforms not only in extraction but also in grouping these information accurately. Fig.~\ref{fig:LLM_KIEval} shows an example where GPT-4o correctly extracts key information but groups it into an incorrect structure, showcasing KIEval's utility in offering a new perspective for assessing LLMs in KIE.

\subsection{Metrics from the Correction Cost Perspective}

\begin{table}[t]
\centering
\begin{tabular}{lcccccccc}
\toprule
\multirow{2}{*}{} & \multicolumn{3}{c}{LayoutXLM} & \multicolumn{3}{c}{LayoutLMv3} & \multicolumn{2}{c}{Donut} \\
 & SROIE & \multicolumn{1}{c}{CORD} & \multicolumn{1}{c}{FUNSD} & SROIE & \multicolumn{1}{c}{CORD} & \multicolumn{1}{c}{FUNSD} & SROIE & \multicolumn{1}{c}{CORD} \\
\midrule
FP + FN & 231 & 190 & 637 & 226 & 218 & 566 & 447 & 404 \\
\midrule
Subs & 93 & 37 & 198 & 102 & 53 & 142 & 219 & 124 \\
Add & 7 & 59 & 126 & 9 & 56 & 136 & 9 & 78 \\
Del & 38 & 57 & 115 & 13 & 56 & 146 & 0 & 78 \\
\midrule
Correction & 138 & 153 & 439 & 124 & 165 & 424 & 228 & 280 \\
\bottomrule
\end{tabular}
\caption{Comparison of FP + FN and Correction (Subs + Add + Del) statistics. Both Add and Del indicate the sum of counts within the matched and unmatched groups. Note: Correction refers to the number of correction steps taken.}
\label{tbl:match_stats}
\end{table}

As previously discussed in Fig.~\ref{fig:problem_definition_FPFN}, the disparity in the conceptualization of KIE errors (either as \{FP, FN\} or as correction cost) results in assessment of KIE models that is misaligned with industrial applications. Table~\ref{tbl:match_stats} shows distinctive gap between the $FP+FN$ and Correction Cost values consistent across all models and datasets. Our work, brings to light of this discrepancy and proposes KIEval$_{\text{Aligned}}$ formulation to better align KIE evaluation to application settings.

\section{KIE Evaluation for RPA System}

\begin{figure}[t]
    \centering
    \includegraphics[width=0.5\linewidth]{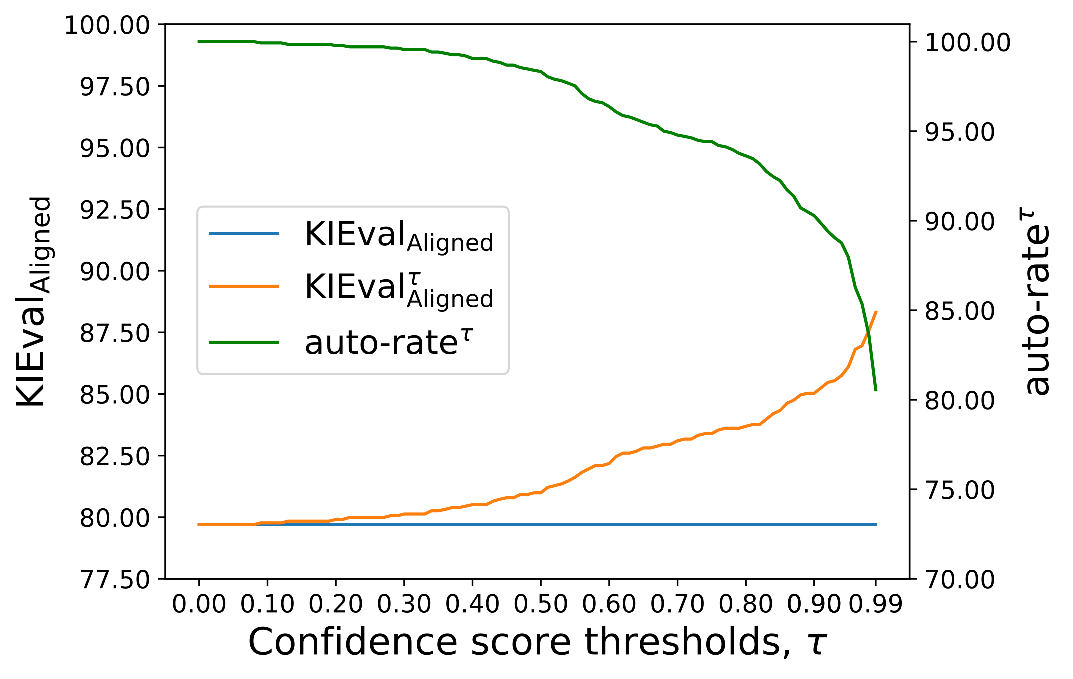}
    \caption{In CORD, Donut's KIEval$_{\text{Aligned}}$  and KIEval$_{\text{Aligned}}^{\tau}$ in relation to varying confidence score thresholds, $\tau$, alongside the corresponding automation rates, auto-rate$^{\tau}$. Increase in confidence score thresholds leads to an increase in KIEval$_{\text{Aligned}}^{\tau}$, while the automation rate decreases due to the rising number of entities requiring human revision.}
    \label{fig:conf_th_graph}
\end{figure}

In addition to the inclusion of structural relation and alignment of metric formulation, there exists a distinctive factor of human-correction that warrants attention when evaluating KIE models in RPA systems. Irrespective of the KIE model's training, it is improbable to consistently achieve error-free extraction performance across a diverse range of documents. In view of this improbability, human-correction (correction by human-intervention) is commonly adopted by RPA systems. Human-correction however, requires a method for selecting a subset of predictions, as verifying and correcting all extracted information is impractical and undermines the very goal of automation in RPA.

Existing RPA systems commonly adopt confidence score based correction where information extracted with confidence below a specific threshold (i.e. uncertain) is selected for verification and correction (if necessary). Selection of optimal threshold value is an application-specific decision that differs from one RPA system to another, contingent on the system's inclinations to trade-off automation rate for KIE performance. In this work, we demonstrate this trade-off analysis with KIEval and highlighting its added insights over prior metrics.

We propose a method to analyse this trade-off in terms of post-correction KIE performance, automation rate, and confidence score threshold value, $\tau$. We first define the automation rate, auto-rate$^{\tau}$, which reflects the proportion of model predictions processed without human verification, interpreted as the number of entities with a confidence score higher than $\tau$. Post-correction KIE performance, KIEval$_{\text{Aligned}}^\tau$, denotes the final KIE performance after KIE predictions with confidence scores below $\tau$ are verified and corrected by humans. A formal definition of these two formulations is provided in the Appendix C.

Fig.~\ref{fig:conf_th_graph} presents the auto-rate$^{\tau}$ and KIEval$_{\text{Aligned}}^\tau$ as a function of the confidence score threshold, $\tau$ in Donut's performance on CORD. The trade-off trend depicted in Fig.~\ref{fig:conf_th_graph} indicates that, as the threshold value increases, the number of information extracted requiring human review increases, leading to a higher post-correction KIE score at the cost of reduced automation rate. Incorporating such trade-off analysis in evaluation of KIE models not only provides deeper insights but also enables stakeholders to conduct cost-benefit evaluations effectively and determine the optimal threshold value for their RPA system.

\section{Conclusion}

In this work, we bring to light of the discrepancies between the existing Document KIE evaluation metrics and the key consideration factors of industrial settings, such as RPA systems. We identify the challenges behind these discrepancies and propose KIEval, metric formulated with an application-centric design. Specifically, KIEval leverages group matching data between the predictions and ground-truth groupings to integrate structural relations in KIE evaluations, differentiating itself from prior metrics that lack grouping awareness in evaluation. Additionally, KIEval formulates KIE errors in terms of the corrections incurred in automation systems (i.e. Substitution, Addition, or Deletion) further bridging the gap between the evaluation metric and industrial settings. The experiments not only verify these discrepancies in existing metrics but also shows how KIEval provides a different perspective of KIE model evaluation from the industrial application's standpoint. On top of these discrepancies, we also demonstrate an application use-case scenario that illustrates the valuable insights which the trade-off analysis brings to RPA systems. This aspect has been overlooked in prior Document KIE metrics. We believe that KIEval could serve as a standard evaluation metric for various KIE tasks and encourage the research community to focus on solving the remaining challenges in KIE tasks with the industrial application in mind.

\bibliographystyle{splncs04}
\bibliography{references}

\title{Supplementary - KIEval: Evaluation Metric\\for Document Key Information Extraction}
\author{Minsoo Khang \and
Sang Chul Jung \and
Sungrae Park \and
Teakgyu Hong}
\institute{Upstage AI, South Korea \\ 
\email{\{mkhang, eric, sungrae.park, tghong\}@upstage.ai}}

\maketitle              %

\appendix

\section{Datasets}

The following table provides details on the datasets and models used in the experiments conducted in this paper. All datasets and models are available on Hugging Face to ensure experimental reproducibility.

\begin{table}[h]
\centering
\begin{tabular}{lccc}
\toprule
       & LayoutXLM                & LayoutLMv3                & Donut                  \\
\midrule
Models & microsoft/layoutxlm-base & microsoft/layoutlmv3-base & naver-clova-ix/donut-base                      \\
\midrule
SROIE  & \multicolumn{2}{c}{darentang/sroie}                  & podbilabs/sroie-donut  \\
CORD   & \multicolumn{2}{c}{nielsr/cord-layoutlmv3}           & naver-clova-ix/cord-v2 \\
FUNSD  & \multicolumn{2}{c}{nielsr/funsd-layoutlmv3}          & -                   \\
\bottomrule
\end{tabular}
\caption{Hugging Face Models and Datasets used in the experiments.}
\label{tbl:hf_models_datasets}
\end{table}

\section{Multimodal LLM prompt for CORD KIE}
Following prompt is used when experimenting with GPT-4o, Qwen2-VL and InternVL 2.5 for KIE in the CORD dataset. Arrow symbol $\hookrightarrow$ represents new-line wrapping in the following text.
\begin{Verbatim}[breaklines=true]
You will be provided with a receipt as an image.
Your task is to analyze the receipt carefully and extract key
information from it.
The entities to be extracted along with their descriptions are
provided below.

| Category | Sub-Category (if applicable) | Entity | Description |
| --- | --- | --- | --- |
| menu | (not applicable) | menu.cnt | quantity of menu |
|      | (not applicable) | menu.discountprice | discounted price of menu |
|      | (not applicable) | menu.etc | others |
|      | (not applicable) | menu.itemsubtotal | price of each menu after discount applied |
|      | (not applicable) | menu.nm | name of menu |
|      | (not applicable) | menu.num | identification # of menu |
|      | (not applicable) | menu.price | total price of menu |
|      | sub | menu.sub_cnt | quantity of submenu |
|      | sub | menu.sub_nm | name of submenu |
|      | sub | menu.sub_price | total price of submenu |
|      | sub | menu.sub_unitprice | unit price of submenu |
|      | (not applicable) | menu.unitprice | unit price of menu |
|      | (not applicable) | menu.vatyn | whether the price includes tax or not |
| sub_total |  (not applicable) | sub_total.discount_price | discounted price in total |
|          |  (not applicable) | sub_total.etc | others |
|          |  (not applicable) | sub_total.service_price | service charge |
|          |  (not applicable) | sub_total.subtotal_price | subtotal price |
|          |  (not applicable) | sub_total.tax_price | tax amount |
| total | (not applicable) | total.cashprice | amount of price paid in cash |
|       | (not applicable) | total.changeprice | amount of change in cash |
|       | (not applicable) | total.creditcardprice | amount of price paid in credit/debit card |
|       | (not applicable) | total.emoneyprice | amount of price paid in emoney, point |
|       | (not applicable) | total.menuqty_cnt | total count of quantity |
|       | (not applicable) | total.menutype_cnt | total count of type of menu |
|       | (not applicable) | total.total_etc | others |
|       | (not applicable) | total.total_price | total price |

Each entity (e.g. menu.cnt) is part of a category (e.g. menu).
You are to extract the entities from the receipt and return in the following format:
```json
{{
    "menu": <dictionary or list of dictionaries>,
    "sub_total": <dictionary or list of dictionaries>,
    "total": <dictionary or list of dictionaries>
}}
```

Note the following characteristics:
1. All entities falling under the same category should be grouped together (represented as a dictionary, such as {total.cashprice, total.changeprice, ...}).
2. If there are multiple entities of the same category, they should be represented as a list of dictionaries.
3. If an entity is not present in the receipt, it should be excluded from the dictionary.
4. Each of the entity's value should either be a string or a list of strings.
5. Note that menu.sub represents a sub-category of the menu category. As such, all entities under menu.sub should be grouped together (either dictionary or list of dictionaries) under the same menu group.
6. You are to respond in JSON format only and ensure that the keys in the dictionary are exactly the same as the entities provided above.
7. If you are unable to extract any information, please return an empty list for that category.

Here is an example of the expected return format:
```json
{
  "menu": [
    {
      "menu.nm": "SPGTHY BOLOGNASE",
      "menu.cnt": "1",
      "menu.price": "58,000"
    },
    {
      "menu.nm": "PEPPER AUS",
      "menu.cnt": "1",
      "menu.price": "165,000",
      "menu.sub": {
        "menu.sub_nm": "WELL DONE"
      }
    },
    {
      "menu.nm": "WAGYU RIBEYE",
      "menu.cnt": "1",
      "menu.price": "195,000",
      "menu.sub": {
        "menu.sub_nm": "MEDIUM WELL"
      }
    }
  ],
  "sub_total": {
    "sub_total.subtotal_price": "503,000",
    "sub_total.service_price": "25,150",
    "sub_total.tax_price": "52,815"
  },
  "total": {
    "total.total_price": "580,965"
  }
}
```
\end{Verbatim}

\section{Automation Trade-off Analysis Metric}

Prior metrics, including $\text{KIEval}_{\text{Aligned}}$ defined above, evaluate Document KIE models without consideration of the full pipeline of Document KIE applications. The RPA system commonly employs a human-correction stage after model inference. Specifically, the RPA system utilizes confidence scores of the extracted entities by a Document KIE model and identifies which entities require further manual verification and corrections with a certain threshold, $\tau$, of the confidence score. We assume that human correction is only conducted on the predictions with lower confidence scores and considers only substitution and deletion without any addition operations because addition operation usually requires examining all predictions and ground-truths, making the correction process and the RPA system inefficient. 

To illustrate the formulation, let $c(\text{pr}_{n,i})$ be the confidence score of $\text{pr}_{n,i}$, where $\text{pr}_{n,i}$ indicates the $i$-th entity in the $n$-th predicted group. $\mathbf{PR}^{<\tau}$ is the set of the predictions of which confidence scores are less than the threshold $\tau$. Since $\mathbf{PR}^{<\tau}$ is only reviewed among the total $\mathbf{PR}$, the automation rate of the RPA system can be defined as follows:
\begin{equation}
    {\text{auto-rate}}^{\tau} = 1 - {|\mathbf{PR}^{<\tau}|} / {|\mathbf{PR}|} \label{eqn:AutomationRate}.
\end{equation}
If the automation rate becomes close to 0 with high $\tau$, the system becomes inefficient but the output of the system becomes accurate. When the automation rate is close to 1 with sufficiently low $\tau$, the system becomes efficient but at the cost of potentially containing incorrect predictions by skipping the human-correction stage. 

To control the trade-off between the system efficiency and accuracy, we introduce KIEval$_{\text{Aligned}}^{\tau}$ that evaluates the accuracy of the RPA automation system with the human-correction stage. The evaluation assumes no human error in the correction stage and the errors in $\mathbf{PR}^{<\tau}$ are only revised with substitution and deletion operations. After the correction process, the remaining errors can be categorized into $\text{Subs}^{\tau}$, $\text{Del}^{\tau}$, and $\text{Add}$. $\text{Subs}^{\tau}$ and $\text{Del}^{\tau}$ denote the error present in predictions with confidence score higher than $\tau$, while $\text{Add}$ represents the number of required entities missed in $\mathbf{PR}$. With the remaining error counts, KIEval$_{\text{Aligned}}^{\tau}$ can be calculated as follows:
\begin{equation}
    \text{KIEval}_{\text{Aligned}}^{\tau} = 1 - \frac{\text{Subs}^{\tau} + \text{Del}^{\tau}+\text{Add}}{\text{N}(\mathbf{PR}^*)+\text{Add}},
\end{equation}
where $\text{N}(\mathbf{PR}^*)$ indicates the number of predictions, $\mathbf{PR}^*$, after the human correction stage. The denominator includes $\text{Add}$ to represent the total number of entities of the system output, including the entities missing in $\mathbf{PR}^*$. Through ${\text{auto-rate}}^{\tau}$ and KIEval$_{\text{Aligned}}^{\tau}$, the automation efficiency and accuracy of the RPA system can be measured by adjusting the confidence threshold $\tau$, facilitating their trade-off analysis in Document KIE.

\end{document}